\documentclass{article}
\usepackage{spconf,amsmath,graphicx}
\usepackage{pgfplots}
\usepackage{pgfplots, pgfplotstable}
\usepackage{numprint}
\pgfplotsset{compat=1.11,
        /pgfplots/ybar legend/.style={
        /pgfplots/legend image code/.code={%
        \draw[##1,/tikz/.cd,bar width=3pt,yshift=-0.2em,bar shift=0pt]
                plot coordinates {(0cm,0.8em)};},
},
}
\pgfplotsset{compat=1.14}
\usetikzlibrary{patterns}

\usepackage{todonotes}

\title{Retrieving Speaker Information from Personalized  Acoustic Models for Speech Recognition}
\name{Salima Mdhaffar$^1$, Jean-François Bonastre$^1$, Marc Tommasi$^2$, Natalia Tomashenko$^1$, Yannick Estève$^1$ \thanks{This work was supported by the French National Research Agency under project DEEP-PRIVACY (ANR-18-CE23-0018).}}
\address{$^1$ LIA, Avignon Université, France \\
$^2$ Université de Lille, CNRS, Inria, Centrale Lille, UMR 9189 - CRIStAL, Lille, France}
\begin{document}
%
\maketitle
\begin{abstract}

The widespread of powerful personal devices capable of collecting voice of their users has opened the opportunity to build speaker adapted speech recognition system (ASR) or to participate to collaborative learning of ASR. In both cases, personalized acoustic models (AM), i.e. fine-tuned AM with specific speaker data, can be built. A question that naturally arises is whether the dissemination of personalized acoustic models can leak personal information. In this paper, we show that it is possible to retrieve the gender of the speaker, but also his identity, by just exploiting the weight matrix changes of a neural acoustic model locally adapted to this speaker. Incidentally we observe phenomena that may be useful towards explainability of deep neural networks in the context of speech processing.  Gender can be identified almost surely using only the first layers and  speaker verification performs well when using middle-up layers.
Our experimental study on the TED-LIUM 3 dataset with HMM/TDNN models shows an accuracy of 95\% for gender detection, and an Equal Error Rate of 9.07\% for a speaker verification task by only exploiting the weights from personalized models that could be exchanged instead of user data.


\end{abstract}
\section{Introduction}
\label{sec:intro}

Automatic speech recognition (ASR) is now at the heart of a large number of applications used on a daily basis by a large number of users.
In order to improve the performance of their ASR solutions, it is common that companies collect and centralize data to train new acoustic models.
New data regulations such as the General Data Protection Regulation in the European Union change the rules in order to protect the citizen privacy~\cite{nautsch2019gdpr}.
In order to improve the performance of ASR models by leveraging user experience without accessing their data, solutions such as federated learning are increasingly being proposed. They consist on exchanging personalized models, or their gradients, instead of data~\cite{leroy2019federated,hard2020training,guliani2021training,yu2021federated,cui2021federated} to preserve the user privacy. 
In the framework of collaborative distributed learning, a personalized model is a model that has been locally adapted to a user~\cite{PersonalizationMansour2020}.
In a very recent work~\cite{mdhaffar2021study}, we presented a such approach to personalize an hybrid HMM/TDNN acoustic model~\cite{TDNN_PeddintiPK_IS15} in a context of collaborative learning.

In this paper we investigate the information contained in personalized acoustic models.
Especially, we are interested in the information related to the speaker identity or the speaker gender that is retrievable from personalized acoustic models.
Previous works have studied speech intermediate representations computed within neural end-to-end models for speech recognition. They illustrated the way such end-to-end models build phonetic and graphemic representations~\cite{belinkov2017analyzing,belinkov19_interspeech}, or showed how speaker variability and noise are gradually removed as the layer goes deeper~\cite{li2020does}.
To our knowledge, there is no study on the information provided by the changes in neural weight due to an acoustic model personalization. 
Our assumption is that the changes applied to the weights of a neural acoustic model when this model is adapted to a speaker brings information about this speaker. 
Thanks to our experimental protocol we expect to evaluate the level of speaker information that can be retrieved directly from these weight changes, and also highlight in which neural layers these changes are particularly discriminant for such information.

Section~\ref{sec:personalization} describes the acoustic model personalization, section~\ref{sec:information_retrieval} presents the approach proposed to retrieve gender and speaker information from the personalized acoustic models, while section~\ref{sec:experiment} describes the experimental set up and section~\ref{sec:results} the experimental results.

\section{Acoustic model personalization}
\label{sec:personalization}
\begin{figure}[h!]
    \centering
     \includegraphics[scale=0.5]{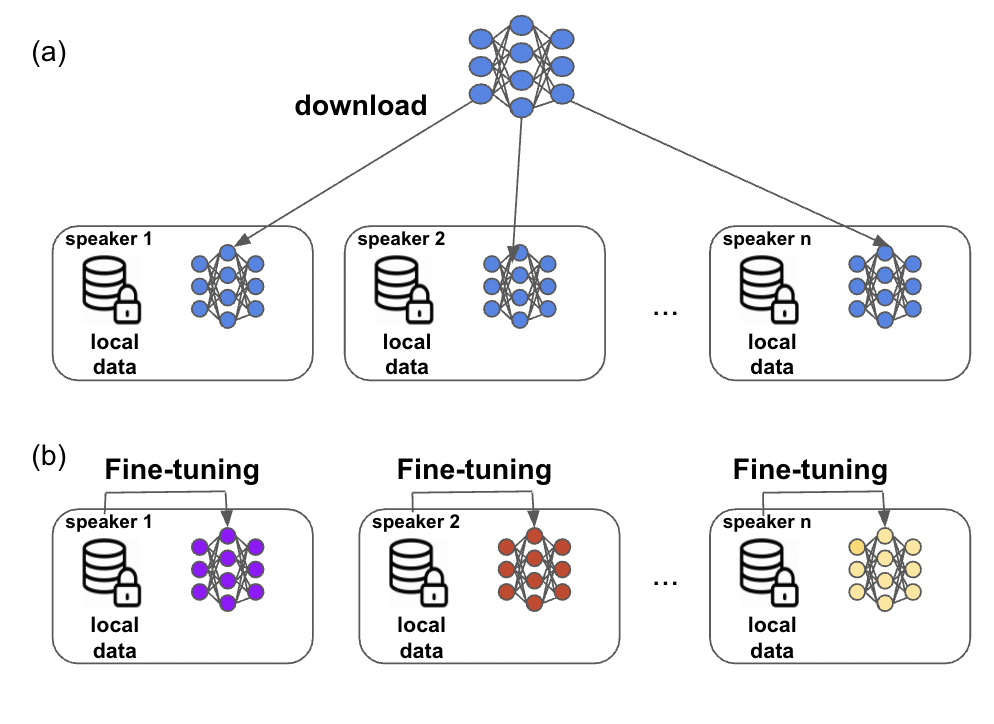}
     \caption{Model personalization: (a) A generic model is downloaded to each user device. (b) The generic model is locally fine-tuned on the user data stored on the device.}
     \label{fig:model_personalization}
\end{figure}

In our scenario, a global acoustic model is available, trained on an initial public dataset.
This global model is distributed to many devices -- each device is linked to only one speaker -- on which it is possible to fine-tune a local instance of the global model by locally exploiting the user data.
Fine-tuning consists in continuing the training process of the generic acoustic model on a small dataset of the target speaker, by taking care on avoiding overfitting.
The output of the fine-tuning process is considered as a personalized model for the local speaker.  
Figure \ref{fig:model_personalization} illustrates the model personalization explored in this work. 
Used in the context of a collaborative distributed learning, for instance federated learning, such personalized models, or their gradients, would be exchanged in order to aggregate and improve a global model without sharing user data in an iterative way.

Even if local data are not transmitted, a possible leakage of personal information appears when personalized models are exchanged. Therefore, our study explores the amount of personal information that can be inferred from the model weights of the personalized models.

\section{Speaker information retrieval}
\label{sec:information_retrieval}

During the model personalization, the weights of the neural generic model are updated.  We assume that these weight updates are dependent to some speaker characteristics.  We expect that it is possible to extract such speaker information by only studying these weight changes. 
We focus on gender and on speaker identity. 
In addition, we investigate in which hidden layers these changes are particularly informative to retrieve such speaker information.

\subsection{Gender information}
In order to retrieve the gender information from the personalized models, we propose an approach based on  clustering into 2 classes. We assume the two clusters corresponds to a female/male classification. 
We evaluate this hypothesis by calculating the purity criterion, by using the gender labels as ground truth. 

We perform as many clustering as the number of layers in the models. 
The inputs are the weights of the layers at the same depth. The clustering algorithm is an agglomerative clustering that merges the closest pair of clusters recursively, building a hierarchy of clusters in bottom-up fashion. The distance between layers is the Euclidean distance and the Ward linkage function is used to evaluate the distance between clusters. It is based on the minimum variance method and allows to minimize the total within cluster variance.


\subsection{Speaker Identification}
In the second part of this study, we want to evaluate the ability to identify speakers, again by only considering the changes applied to weight matrices during the personalization of an ASR acoustic model. 
However, such weight matrices, and even their hidden layers, are too large to characterize the speaker. 
Reduction dimensionality approaches like Principal Component Analysis (PCA) are a potential solution but the large reduction factor targeted, combined to a limited number of samples -- one model by speaker --   could result in this case in a large loss of discriminant information. In order to solve this dimension/discrimination problem, we propose to apply a method inspired from~\cite{snyder2018x}, that consists in learning a speaker embedding extractor.  This neural network-based extractor is trained on the weight matrices of a given hidden layer from personalized neural ASR models. Training objective is a speaker discrimination task. But we have to face two difficulties: the input matrices are very large, and the training dataset is very small.

\begin{figure*}[h!]
    \centering
     \includegraphics[scale=0.35]{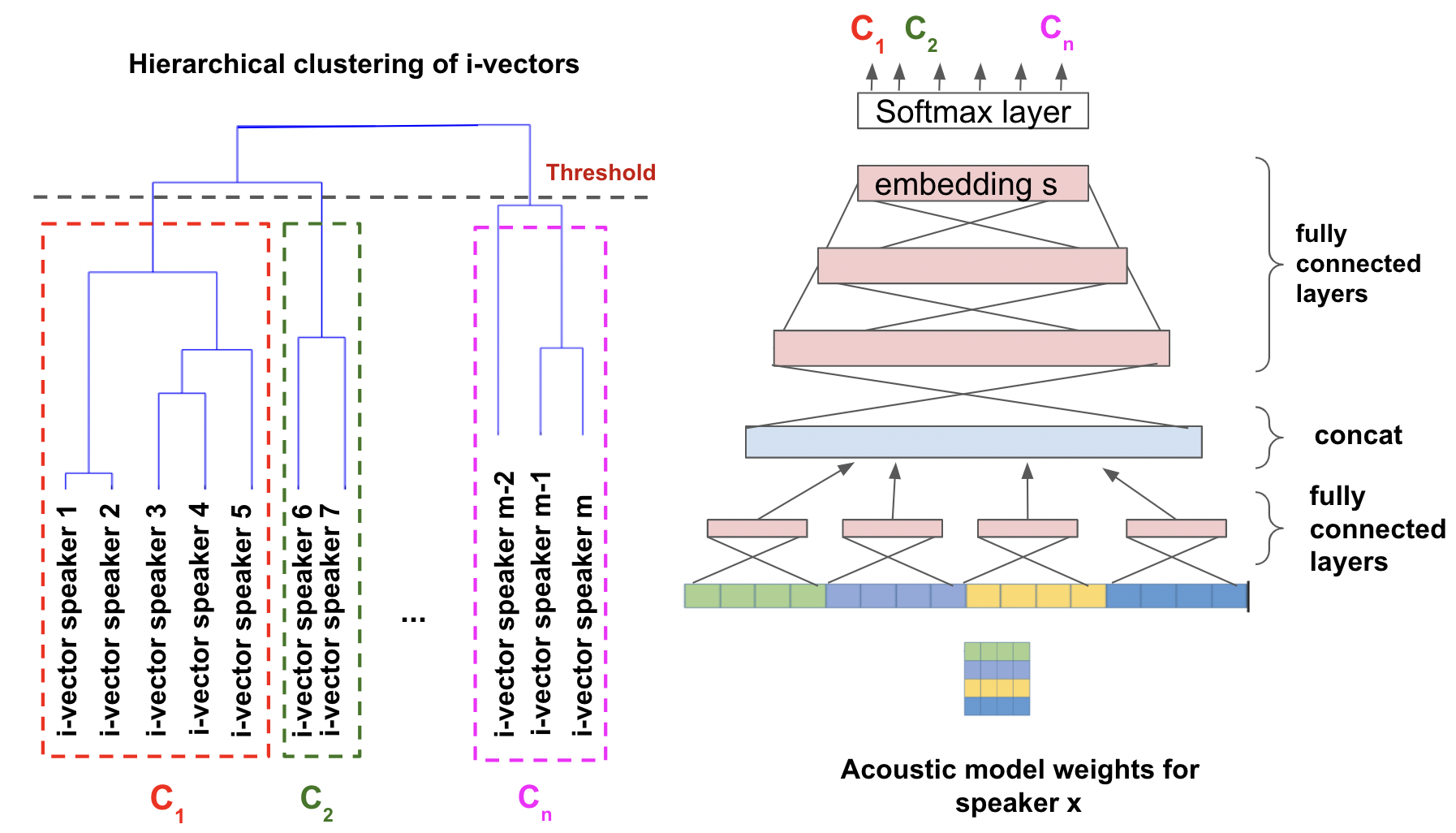}
     \caption{Proposed DNN structure to train speaker's embedding from neural network weights of acoustic models}
     \label{fig:Neural_network_architecture}
\end{figure*}

We propose to modify the training objective by using classes of speakers as classification labels, in place of speakers.  This allows us to increase the number of examples per class during the training phase, and so to reduce  the risk of overfitting. The classes of speakers used for the speaker embedding extractor training are issued from  hierarchical clustering of i-vector  of speakers present in the training data.

In order to drastically reduce the memory footprint of the extractor and overcome this difficulty, we designed a specific structure for our extractor. Starting from a classical deep neural network (DNN) classifier, we apply a multi-stream input approach. The weight matrix is split into small blocks that are separately linked to a dedicated hidden layer. A small block of the input weight matrix is composed by all the weights related to a unit neural in the hidden layer targeted in the ASR acoustic model. 
For instance, if the targeted hidden layer $H_t$ of the ASR acoustic model architecture contains $n$ units, the weight matrix used as input of our speaker embedding extractor will be split into $n$ different blocks.
Next, the outputs of the hidden layer dedicated to each block are concatenated  to feed the upper hidden layer of the DNN-based extractor, composed of fully connected layers followed by the final softmax layer.

The structure of the resulting embedding extractor is illustrated in Figure~\ref{fig:Neural_network_architecture}. The embedding layer is the hidden layer just below the softmax one. The resulting DNN model is able to extract speaker embeddings from speech data, including for speakers that were not present in the training.

\section{Experimental framework}
\label{sec:experiment}
\subsection{ASR system}

Our experiments target (chain) HMM/TDNN acoustic models for speech recognition~\cite{povey2016purely}. 
The ASR system is based on the Kaldi toolkit~\cite{povey2011kaldi}. 
The chain-TDNN setup is based on 13 layers with 512 dimensions and is trained on cepstral mean and variance normalized 40-dimensional MFCC features.
I-vectors are also incorporated as auxiliary input features.
The resulting context of TDNN models is 28 left and 28 right neighbour frames.
The acoustic model has about 14 million parameters. 
For the generic model, the initial and final scheduled learning rates are equal to 0.00025 and 0.000025 respectively. 
Training audio samples are randomly perturbed in speed and volume during the training process. 


\subsection{Dataset}
\label{sect:dataset}
All experiments to train generic and fine-tune acoustic models are conducted with the TED-LIUM~3 dataset~\cite{hernandez2018ted}, a large corpus of 452 hours of TED talks pronounced by 2,295 speakers. 
For the study presented in this paper, an original setup has to be defined.
Similarly to~\cite{mdhaffar2021study}, the dataset  is splited into three parts and the sets of speakers in each part are pairwise disjoints. 
Characteristics of the three parts are reported in Table~\ref{tab:datasetsplit}. 
The first part is called \emph{generic} and has been used to train the initial acoustic model for ASR. 
The two other parts, called \textit{p1} and \textit{p2}, are used for model personalization and evaluation.
In both subsets, the available audio material is split in order to get two sessions of 5 minutes per speaker. 
Each session is used to personalize a model.

Table \ref{tab:datasetsplit} presents the statistics of the three subsets. 
For \textit{p1} and \textit{p2}, the table presents the exact number of speakers who have pronounced enough speech to have two sessions of five minutes (463 from 650 speakers for \textit{p1} and 581 from 765 speakers for \textit{p2}, respectively).



The TED-LIUM~3 dataset is provided without information about the gender.
Using the website of TED conference, the annotation of the corpus in gender was done manually for $\textit{p2}$.\footnote{For the
reproductibility of experimental results by research community, we will make available this annotation.}
Table~\ref{tab:datasetsplit} presents also statistics about the gender.

\begin{table}[h]
  \centering
  \begin{tabular}{|l|l|l|l|}
    \hline
    & generic  &  $p1$ & $p2$ \\ \hline\hline
    Duration (hours) &    200  &  150  &     170 \\ 
    Duration of speech (hours)      &    170   & 125  &   150 \\
    \# speakers  &     880  & 650  & 765 \\ 
    \# speakers (duration$>$10 min)  &     -  & 463  & 581 \\ 
    \# men  &   -    &  - & 553 \\
    \# women  &  -     &  - & 212 \\\hline
  \end{tabular}
  \caption{TED-LIUM 3 dataset}
  \label{tab:datasetsplit}
\end{table}

\subsection{Personalized models}
The initial generic model is trained on the \textit{generic} part.
Personalized models are obtained by fine-tuning the generic model on the speaker's data from \textit{p1} and \textit{p2}: for each speaker, we personalize the generic model twice using separately his/her two five-minutes sessions. 
Thus, for most of the speakers (speakers with duration $>$ 10 minutes), two different personalized models are obtained.

When fine-tuning the generic model on target speaker data, we modify only the value of learning rate (the initial and final learning rates were equal to 0.000025 and 0.000015 respectively) and all hyperparameters (i.e. learning rate and local epochs number) are assumed to be homogeneous among all workers.
\vspace{-0.5em}
\section{Results and analysis}
\label{sec:results}
\subsection{Gender identification}

There are several methods used to evaluate clustering performance. 
In our study, we use the purity.
Purity focuses only on maximising the
total number of true positive responses per cluster. 
Purity values
range between 0 and 1 (perfect clustering).
It is defined as $Purity = \frac{1}{N} \sum_{i=1}^{k} max_{j} |c_{i} \cap t_{j}|$ where $N$ is the number of speakers, $k$ is the number of clusters, $c_{i}$ is a cluster and $t_{j}$ is the classification count for cluster $c_{i}$. 

\begin{figure}[h!]
\centering
\hspace*{-0.32cm}
\begin{tikzpicture}[font=\scriptsize]
    \begin{axis}[ybar,
    height=3.7cm,width=9.3cm,
    ymin=0,ymax=1.2,
    ylabel style = {align=center},
    ylabel={Purity},
    xticklabel style={rotate=90},
    bar shift=0pt,
    xtick=data,
    symbolic x coords={layer 1,layer 2,layer 3,layer 4,layer 5,layer 6,layer 7,layer 8,layer 9,layer 10,layer 11,layer 12,layer 13},
    nodes near coords, 
    nodes near coords align={vertical},
    every node near coord/.append,
    every node/.style={font=\scriptsize},
    every tick label/.append style={font=\small}
    ]
    \addplot[pink!20!black,fill=pink!80!white] table [x=x,y=ave,col sep=comma]{data_gender.csv};
    
    \end{axis}

\end{tikzpicture}
\caption{Clustering purity of  hidden layer weights of the acoustic model using women and men labels as a reference.}
\label{fig:Purity}
\end{figure}
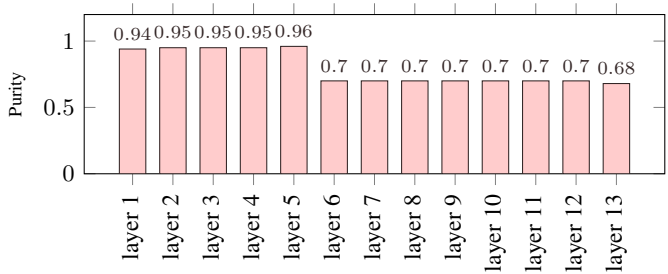

Figure~\ref{fig:Purity} shows the results for the different hidden layers of neural network of ASR acoustic models from which we extract weights for data in \textit{p2}. 
We observe that it is possible to get two gender-based clusters with a purity value of 0.96 for the layer 5.
Results show that gender information can be identified for the five first layers. 

\vspace{-2em}
\subsection{Speaker verification}

Speaker verification is evaluated in terms of false alarm (FA) and false reject (FR) error rates and reported using equal error rate (EER), with  $\mathrm{EER}=\mathrm{FA}=\mathrm{FR}$.

\pgfplotstableread[col sep=comma,]{data22.csv}\datatable

\begin{figure}[h!]
\centering
\begin{tikzpicture}
\begin{axis}[height=4cm,width=9.3cm,
    ybar=3pt,
    bar width=4pt,
    ymax=25,
    xtick=data,
    xticklabels from table={\datatable}{A},
    ylabel={EER (\%)},
    legend pos=north west,
    xticklabel style={rotate=90},
    symbolic x coords={layer 1,layer 2,layer 3,layer 4,layer 5,layer 6,layer 7,layer 8,layer 9,layer 10,layer 11,layer 12,layer 13},
    nodes near coords, 
    every node near coord/.append style={font=\tiny,rotate=90,anchor=west},
    nodes near coords align={vertical},
    every node near coord/.append,
    legend style={at={(0.8,0.9)},
      anchor=north,legend columns=-1, style={font=\small}},
     every tick label/.append style={font=\small},
    legend style={/tikz/every even column/.append style={column sep=0.4cm}}
             ]
    \addplot table [x=A, y=ave]{\datatable};
    \addplot table [x=A, y=p2]{\datatable};
    \legend{p2, p1}
\end{axis}
\end{tikzpicture}
\caption{Speaker verification performance depending on the hidden layer of the acoustic model used to extract weights.}
\label{fig:EER_SV}
\end{figure}
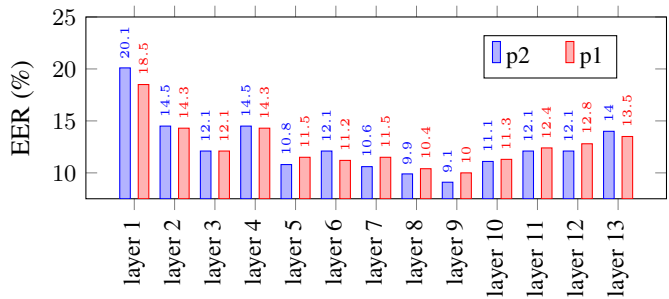

First, a speaker embedding extractor is trained using each layer of our acoustic models as input. To train the extractors, we use 926 personalized  speaker models corresponding to 463+463 unique subsets  from \textit{p1}. 
The trained extractor models are then used on  \textit{p2} data to extract the speaker's embedding (knowing that there is no overlap between \textit{p1} and \textit{p2} speakers).
Respectively, a second experiment is conducted with \textit{p1} as a test set and \textit{p2} as a training set for the extractor.
The number of target classes (issued from a hierarchical clustering of i-vectors of speakers present in the training data) used to train our extractor is fixed to 20 and the dimension of the output vectors, the speaker embeddings, is fixed to 100. 
We use a speaker verification task to evaluate the ability to recognize the speakers from a given layer weights. A simple cosine distance is used to compute the verification score for a trial (enrolment,test).
The data of each speaker (see Section~\ref{sect:dataset}) is divided into two sessions, denoted $s1$ and $s2$. It gives one target trial, $(x_i^{s1},x_i^{s2})$, per speaker $x_i$. Non-target trials, $(x_i^{s1},x_j^{s2})$, are formed by crossing the first session of a given speaker with all the second sessions 
of the other speakers. It gives respectively 463~/~581 target trials and $\numprint{213906}$~/~$\numprint{336980}$ non-target trials for \textit{p1} and \textit{p2}.
Figure~\ref{fig:EER_SV} shows the comparative results in terms of EER.  

The best performance is obtained using layer 9 (9.07\% EER for \textit{p2} and 10\% EER for \textit{p1)}, showing clearly that speaker specific information could be extracted from the weights of a personalised ASR acoustic model.
For comparison purposes, we also computed the performance when the weight vectors are used directly to compute the cosine distance, without the embedding extractor. 
The EER is about 48\% in this case for \textit{p2} (close to the random performance).
This proves the effectiveness of the proposed approach to extract a speaker embeddings from the weights of personalized acoustic model.



\section{Conclusion}
In this study, we showed that it is possible to retrieve the gender and the identity of a speaker from the analysis of the changes applied to the weights of her/his personalized acoustic model.
Experiments conducted on the TED-LIUM 3 dataset show that the gender information is mainly brought by the updates impacting the first five layers of a HMM/TDNN acoustic models composed of 13 hidden layers, when the speaker identity is mainly embedded in the  middle-up hidden layers (5 to 9). To obtain the latter result, we also proposed an original way to build a speaker embedding extractor from personalized weight matrices. 
We obtained a gender purity of 0.96 on the five first layers and a speaker verification EER of ~9\% for layer 9.
These results would be particularly interesting for future works focusing on distributed learning for privacy preservation. In this direction, we propose in a parallel study dedicated to attack approaches against federated learning for speech recognition, 
to use external speech data in order to analyze the behavior of personalized models on such data, see~\cite{tomashenko2022}. 





\bibliographystyle{IEEE}
\bibliography{mybib}

\end{document}